\documentclass[10pt,a4paper,twoside,english]{article}

\usepackage[utf8]{inputenc}
\usepackage[T1]{fontenc}
\usepackage{graphicx,tikz,arydshln}

\usepackage{hyperref}

\usepackage{jeptaln2012}
\usepackage[frenchb]{babel}

\title{Wikipedia Arborification and\\ Stratified Explicit Semantic Analysis}

\author{Yannis Haralambous\up{1}\quad Vitaly Klyuev\up{2}\\[6pt]
{\small  (1) Institut Télécom, Télécom Bretagne \& UMR CNRS 6285 Lab-STICC,
Technopôle Brest Iroise,
CS~83818, 29238 Brest Cedex 3, France\\ 
  (2) University of Aizu,
Aizu-Wakamatsu,
Fukushima-ken 965-8580, Japon 
  \texttt{yannis.haralambous@telecom-bretagne.eu, vkluev@u-aizu.ac.jp} \\ 
}}

\begin{document}

\maketitle

\resume{
Nous présentons une extension du procédé d'analyse sémantique explicite de Gabrilovich et Markovitch. À l'aide de leur mesure de parenté sémantique, nous pondérons le graphe des catégories de Wikipédia. Puis, nous en extrayons un \emph{arbre couvrant minimal} par le biais de l'algorithme de Chu-Liu \& Edmonds. Nous définissons une notion de \emph{tfidf stratifié}, les strates étant, pour une page Wikipédia et un terme donnés, le tfidf classique et les tfidfs catégoriels dans les catégories ancêtres, au sens de l'arbre couvrant minimal. Notre méthode se sert de ce tfidf stratifié, qui favorise les termes qui \og survivent\fg{} lorsque on passe des pages aux catégories, en se dirigeant vers la racine de l'arbre. Nous l'évaluons par une classification de textes tirés du corpus WikiNews, et constatons qu'elle apporte un gain de précision de 18\%. Nous terminons par une série de pistes de recherches futures. 
}

\abstract{[This is the translation of paper \emph{Arborification de Wikipédia et analyse sémantique explicite stratifiée} submitted to TALN 2012.]}{
We present an extension of the Explicit Semantic Analysis method by Gabrilovich and Markovitch. Using their semantic relatedness measure, we weight the Wikipedia categories graph. Then, we extract a \emph{minimal spanning tree}, using Chu-Liu-Edmonds' algorithm. We define a notion of \emph{stratified tfidf} where the stratas, for a given Wikipedia page and a given term, are the classical tfidf and categorical tfidfs of the term in the ancestor categories of the page (ancestors in the sense of the minimal spanning tree). Our method is based on this stratified tfidf, which adds extra weight to terms that ``survive'' when climbing up the category tree. We evaluate our method by a text classification on the WikiNews corpus: it increases precision by 18\%. Finally, we provide hints for future research\rule[-5pt]{0pt}{0pt}.}

\motsClefs{Mesure de parenté sémantique, Wikipédia, analyse sémantique explicite, tfidf catégoriel, tfidf stratifié, arborification, algorithme de Chu-Liu \& Edmonds, classification de texte}
{Semantic relatedness measure, Wikipedia, Explicit Semantic Analysis, category tfidf, stratified tfidf, arborification, Chu-Liu \& Edmonds algorithm, text categorization}

\pagebreak
\selectlanguage{english}

[This is the translation of paper \emph{Arborification de Wikipédia et analyse sémantique explicite stra\-ti\-fiée} submitted to TALN 2012.]

\section{Introduction}

\subsection{Explicit Semantic Analysis}

Unlike semantic similarity measures, which are limited to ontological relations such as synonymy, hyponymy, meronymy, etc., \emph {semantic relatedness} measures detect and quantify semantic relations of a more general kind. The typical example is the one of concepts \textsc {car}, \textsc {vehicle} and \textsc {gasoline}. A car is a special kind of a vehicle, so we have an hyperonym relation and this can easily be measured by a semantic similarity measure (for example, by taking the inverse of the length of the shortest path between these concepts in WordNet). But between \textsc {car} and \textsc {gazoline}, there is no semantic similarity, since a car is a solid object and fuel is a liquid. But there is an obvious semantic relation between them since most cars use gasoline as their energy source, and this relation will be measured by the semantic relatedness measure.

\cite {Gabrilovich:2007uo} introduce a semantic relatedness measure called ESA (=~Explicit Semantic Analysis), as opposed to the popular method of \emph{latent semantic analysis} \cite {lsa-en}. ESA is based on the Wikipedia corpus. The principle is simple: after cleaning and filtering Wikipedia pages (keeping only those with a sufficient amount of text and a given number of incoming and outgoing links), they remove stop words, stem all words and calculate their tfidfs. A Wikipedia page can then be represented as a vector in the space of (nonempty, stemmed, distinct) words; the coordinates of the vector are the tfidf values (normalized so as to have unit vectors and thus remain independent of the size of the page). 

By the very nature of Wikipedia, we can consider that every page is a concept. All the concepts thus form a matrix whose columns are concepts and whose lines are words. By transposing this matrix we obtain a representation of words in the space of concepts. The ESA measure of two words is simply the cosine of their vectors in this space.

Intuitively, two words are ESA-semantically close if they appear frequently in the same Wikipedia pages (so that their tfs are high), and rarely in the whole corpus (for their dfs to be low). 

Despite the good results obtained by this method, it has given rise to some criticisms. Thus, in \cite {Haralambous:2011wm} we notice that ESA has poor performance when the relation between words is mainly ontological. As an example, the word ``mile'' (length unit) does not appear in the page of the word ``kilometer'' and the latter appears only once in the page of the former: that is hardly sufficient to establish a nonzero semantic relatedness value; however, such a relation is obvious, since both words refer to units of length measurement. An ontological component, obtained from a WordNet-based measure can fill this gap.

Another, more fundamental criticism is that of \cite {Gottron:2011im}, who argue that the choice of Wikipedia is irrelevant, and that any corpus of comparable size would give the same results. To prove it, they base ESA not on Wikipedia, but on a Reuters news corpus, and get results even better than with standard ESA. According to the authors, the semantic relatedness value depends only on the collocational frequency of the terms, and this whether documents correspond to concepts or not. In other words they deny the ``concept hypothesis,'' saying that ESA uses specifically the correspondence between concepts and Wikipedia pages.

In this article we will enhance ESA by adopting a different approach: the persistence of tfidfs of terms when leaving pages and entering the category graph.

\subsection{Wikipedia Categories}

Wikipedia pages are generally well written and structured, but unfortunately this is not always the case for page \emph {categories}. This is due to two factors: first, the choice of categories  is the result of a collective effort, and this is not always the ideal condition for obtaining a coherent result (not to mention the fact that writing a Wikipedia page requires good knowledge of a given topic while the choice of a category requires good knowledge of Wikipedia \emph{as a whole}, which is less common among wikinauts). Second factor, there is no strict separation between thematic and utilitarian categories. For example, categories are used both to classify pages according to their themes, as to identify pages  being too short or needing corrections or are likely to be deleted, and so on.

Finally, what makes NLP applications more difficult is the fact that categories form a fairly complex graph, which, in particular, contains cycles. Thus, according to \cite {Medelyan:2009ts}, ``cycles are not encouraged but may be tolerated in rare cases.'' The very simple example of categories ``\emph{Zoologie}'' and ``\emph{Animal}'' pointing to each other, shows that the semantic relation underlying subcategories is always hyperonymy. Here \textsc {animal} is the object of study of discipline \textsc {zoologie}. We attempted the following experiment: starting from the 2,782,242 (unfiltered) French Wikipedia pages, we followed paths formed by the category links. The choice of each subsequent category was made at random, but did not change during the experiment. Among these paths, 2,162,115 resulted in cycles. It turned out that it was always the same 50 cycles, 12 of which were of length~3 (triangles) and all others of length~2 (categories pointing at each other, as in the example above, which was detected by this method). It results from our experiment that the problem is real but not insurmountable, since the number of cycles probably can be kept within reasonable limits.

\subsection {Related Work}

\cite {Scholl:2010tu} enhance the performance of ESA using categories. They proceed as follows: let $ T $ be the matrix whose rows represent the Wikipedia pages and whose columns represent words. The value $ t_ {i, j} $ of cell $(i, j) $ is the normalized tfidf of the $ j $th word in the $ i $th page. For each word $ m $ there is therefore a vector $ \vec {v}_m $ whose dimension is equal to the number of pages. Now let $ C $ be the matrix whose columns are pages and whose lines are categories. The value of a cell $ c_ {i, j} $ is 1 when page $ j $ belongs to category $ i $ and 0 otherwise. Then they take the product of matrices $ \vec {v}_m \cdot C $ which provides a vector whose $ j $th component is $ \sum_ {i \mid D_i \in c_j} t_ {i, j} $, that is the sum of tfidfs of word $ m $ for all pages belonging to the $ j $th category. They use the concatenation of  vector $ \vec {v}_m $ and of the transpose of $ \vec {v}_m \cdot C $ to improve system performance on the text classification task. They call this method XESA (eXtended ESA).

We see that Scholl et al. extend page tfidf to categories by simply taking the sum of tfidfs of all pages belonging to a given category. This approach has a disadvantage when it comes to high-level categories: instead of being a way to find the words that  characterizing a given category, the tfidf of a word tends to become nothing more than the average density of the word in the corpus, since for large categories, tf tends to the total number of occurrences of the word in the corpus, while the denominator remains constant and equal to the number of documents containing the given word. Thus, this type of tfidf loses its power of discrimination for high-level categories. We propose another extension of tfidf to categories, which we call \emph {categorical tfidf}. The difference lies in the denominator, where we take, the number not of all documents containing the term, but only of those not belonging to the category. Thus our categorical tfidf (which is equal to the usual tfidf in the case of pages) is high when the term is common in the category and \emph {rare elsewhere} (as opposed to \emph {rare on the entire corpus} of Scholl et al.).

On the other hand, in \cite {Collin:2010}, the authors pose the problem of inconsistency of Wikipedia's category graph and propose a shortest path approach (based on the number of edges) between a page and the category ``\emph{Article},'' which is at the top of the hierarchy. The shortest path provides them with a semantic and thematic hierarchy.

However, as already observed in the case of WordNet \cite [p.~275] {wordnet}, the length (in number of edges) of the shortest path can vary randomly, depending on the density of  pages (concepts, in the case of WordNet) in a given domain of knowledge. On the other hand, the distance (in number of edges) between a leaf and the top of the hierarchy is often quite short, requiring frequently an arbitrary choice between paths of equal length.

What is common with our approach is the intention to simplify Wikipedia's category graph. But instead of taking the number of edges, we weight the graph using ESA measure and utilize that weight for our simplification. This weight, which is based on the statistical presence of words in pages belonging to a given category, allows us to calculate a minimum spanning tree. The result of this operation is that any page (or category, other than ``\emph{Article}'') has exactly one parent category that is semantically closest to it. This happens in a global sense, that is to say, so that the total weight of the tree is minimum.

We use this tree to define a notion of \emph {stratified tfidf}. Our goal is to avoid words which, by chance, have a high tfidf in a given page despite the fact that they do not really belong to the theme of the page. Our hypothesis is that a word having an unduly high tfidf will disappear when we calculate its tfidf in higher categories. Only the words in line with the theme of the page ``survive'' when we move away from leaves of the tree and towards the root.

In this way the ``concept hypothesis,'' denied by \cite {Gottron:2011im}, is again essential: when there is no concept, there is no hierarchy and when there is no hierarchy,  survival of words in categories would be random and not due to the inherent hierarchy of the path. 

\section {Adaptation of ESA to French Wikipedia}

To adapt ESA to French Wikipedia, we followed the same steps as \cite {Gabrilovich:2007uo} and \cite {calli} except for one thing: we have preceded the stemming step by lemmatization, to avoid loss of information due to poor stemming of inflected words (in English this  phenomenon is negligible, so that stemming can be performed directly).

By limiting the minimum size of pages to 125 (nonstop, stemmed and distinct) words, 15 incoming and 15 outgoing links, we obtained a number of Wikipedia pages equivalent to the original ESA method: 128,701 pages (out of 2,782,242 in total) containing 1,446,559 distinct words (only 339,679 of which appear more than three times in the corpus).

\cite {Gabrilovich:2007uo} evaluate their method on WS-353, a set of 352 pairs of English words, the semantic relatedness of which has been evaluated by 15--16 human judges. Their criterion is the Spearman correlation coefficient between the rank of pairs obtained by ESA and the one obtained by taking the average of human judgments. We have translated these pairs into French, but the result was very disappointing. Indeed, some twenty words are untranslatable into a single term (the current version of ESA covers only single-word terms), such as ``seafood'' which can be translated only as ``\emph{fruits de mer}.'' Furthermore there are ambiguities of translation resulting from word polysemy. When we translate the pair ``flight/car'' by ``\emph{vol}/\emph{voiture},'' we obtain a high semantic relatedness due to the criminal sense of ``\emph{vol}'' (=~theft) while the sense of the English word ``flight'' is mainly confined to the domain of aviation. Finally, some obvious collocations disappear when translating word to word, such as ``soap/opera'' which is unfortunately not comparable to ``\emph{savon}/\emph{opéra}''\ldots

We have therefore chosen to evaluate our implementation of ESA in a more traditional way, by performing a text classification task. From the French WikiNews corpus, we have extracted four classes of documents corresponding to the following topics: politics, economics, culture, sports. Here are the characteristics of our test corpus:
\begin {center}
\small \begin {tabular} {|c|c|c|c|} \hline
Class & Number of text & Number of words &Number of ns.s.d. words\\\hline
Politics&466&70,829&5,537\\
Economy&710&97,610&6,951\\
Culture&327&46,812&6,063\\
Sports&1,671&166,174&7,792\\\hline
Total&3,174&381,425&16,323\\\hline
\end{tabular}
\end{center}
where the 4th column contains the number of nonstop stemmed distinct words.

We represent each document as follows: let $ w $ be a word in document $ d $, $ \vec {w} $ the vector of the word in the space of concepts, $t_d  (w) $ its tfidf\footnote {\cite {Gabrilovich:2007uo} define tfidf as $ t_d (w) = (1 + \log (f_d (w))) \cdot \log \left (\frac {\# \mathcal {W}} {\mathrm {df} (w)} \right) $, where $ f_d (w) $ is the frequency of $ w $ in page $d $, $\# \mathcal {W} $ is the total number of Wikipedia pages and $ \mathrm {df} (w) $ is the number of pages containing $ w $.} in $ d $. Then the vector $ \vec {d} $ of document $ d $ defined as\label {vec}:
$$
\vec {d}: = \frac {\sum_ {w \in d} t_d (w) \cdot \vec {w}} {\sqrt {\sum_ {w \in d} t ^ 2_d (w)}}.
$$
where the denominator is used for normalization. In this way, all $ \vec {d} $ are of  (Euclidean) norm~1, and thus belong to the sphere $ \mathbb {S} ^ N $ where $ N $ is the number of Wikipedia pages (in our case: $N = 128,701$).

We applied a linear SVM classifier\label {svm} to the set of these vectors and the corresponding classes, and after a tenfold cross-validation, we obtained the following result:
\begin {center}
\small \begin {tabular} {| c | c | c |} \hline
Dim. of subspace& Number of SVs & Precision \\\hline
103,291 & 665 & 70.08\% \\\hline
\end {tabular}

\end {center}
where the dimension of the subspace of concepts is the size of the smallest subspace that contains the vectors of all documents in our corpus. The precision obtained is rather low, which is probably---and at least partly---due to the small size of texts in the corpus. That said, our goal is not to compare ESA to other classification methods, but to show that our approach improves ESA. So, this result is our starting point and we intend to improve it.

\section {Wikipedia Arborification}

\begin{figure}[t]
\resizebox{\textwidth}{!}{\includegraphics{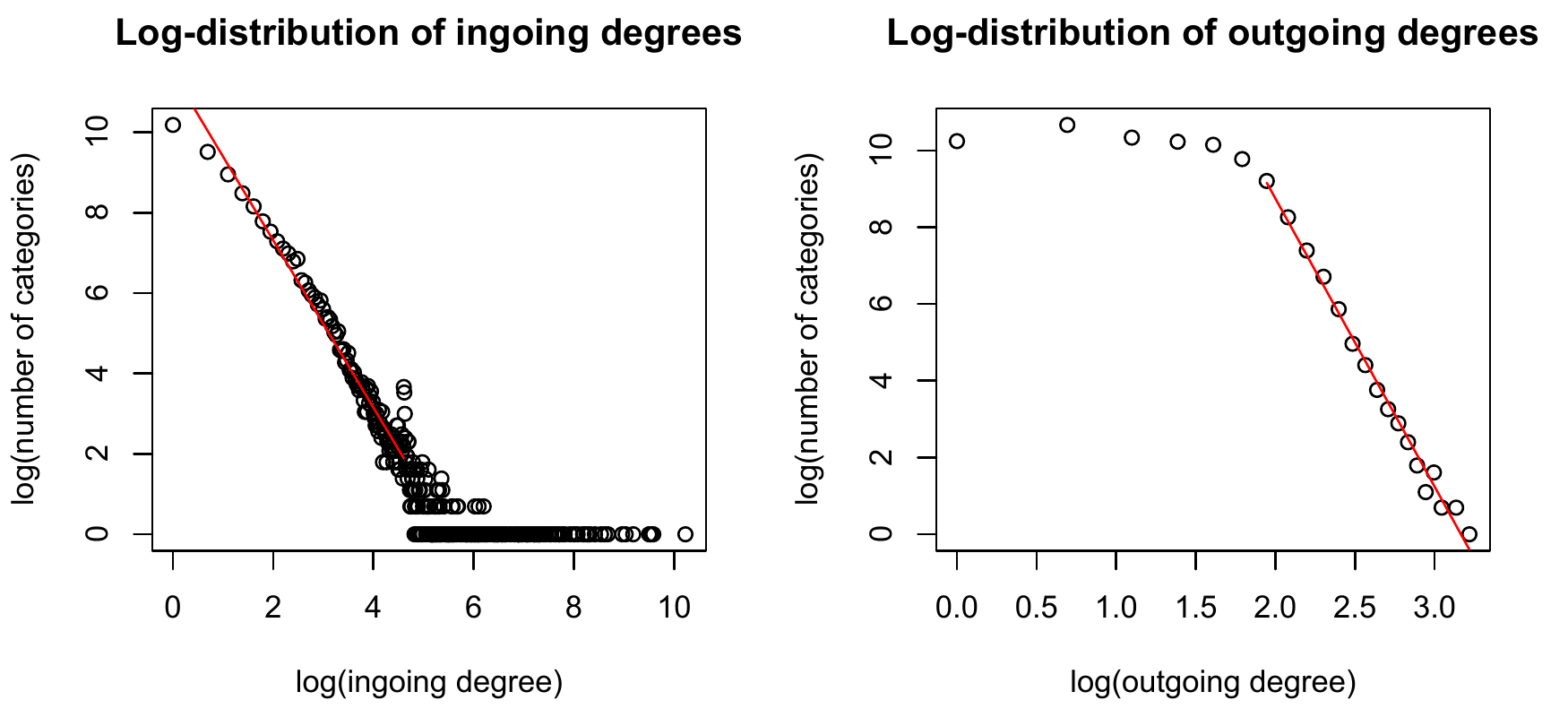}}
\caption{Ingoing and outgoing degree distribution of Wikipedia categories. \label {fig0}}
\end {figure}

Wikipedia's category graph has been studied thoroughly in \cite {zesch}, for the English version of Wikipedia. French Wikipedia (version of December 31, 2011), has 2,782,242 pages, 293,244 categories, 680,912 edges between categories and 12,935,688 edges between pages and categories (note that the same category may have incoming edges from both pages and other categories). As can be seen on Fig.~\ref{fig0}, by the logarithmic distribution of incoming and outgoing degrees, we see that this graph follows a power distribution $ p ^ {- \alpha} $ for $ \alpha =  2.08$ for incoming degrees (a pretty standard value) and $ \alpha =  7.51$ for outgoing degrees (this value is rather high).

These calculations were made on the entire Wikipedia category graph. As explained in the previous section, to calculate ESA we filtered the corpus and selected only the most important pages. This also decreased the number of categories. Furthermore we removed the auxiliary categories (but not ``portals'' since they carry semantics). In the end, remained 104,258 categories for 128,701 pages, with 1,605,946 edges, 387,146 of which were edges between categories.

We first calculated the pages that belong to every category (or to its subcategories), i.e., for a given category $ c $, all pages $ d $ such that there is a membership $ d \to c_1 $ and a sequence of inclusions $ c_1 \to c_2 \to \cdots \to c $. Let $ \mathcal {F} (c) $ be the set of these pages (= the set of ``leaves of $ c $''), $ \mathcal {W} $ the entire (filtered) set of Wikipedia pages and $ \# \mathcal {W} $ its cardinal. If $ w \in d $ is a (nonstop, stemmed, distinct) word of $ d \in \mathcal {W} $ and $ t_d (w) $ its classical tfidf, then we define the \emph {categorical tfidf} $ t_c (w) $ of $ w $ for category~$ c $ as follows:
$$
t_c(w):=\left(1+\log\left(\sum_{d\in\mathcal{F}(c)}f_d(w)\right)\right)\cdot\left(\log\left(\frac{\#\mathcal{W}}{1+\sum_{d\in \mathcal{W}\setminus\mathcal{F}(c)}1}\right)\right).
$$

The difference with the classical tfidf is in the calculation of idf: instead of the $ \sum_ {d \in \mathcal {W}}  1$ used by \cite {Scholl:2010tu}, we focus on the set difference between Wikipedia and leaf pages $ c $, and we use $ 1 + \sum_ {d \in \mathcal {W} \setminus \mathcal {F} (c)}  1$ instead (the unit is added to avoid a zero df when the term does not appear outside the category). We believe that this extension of tfidf to  categories improves discriminatory potential, even when the sets of leaves are large.

Using this tool, we calculated the vectors of categories, defined as follows:
$$
\vec{c}:=\frac{\sum_{w\in \mathcal{F}(c)}t_c(w)\cdot \vec{w}}{\sqrt{\sum_{w\in \mathcal{F}(c)}t^2_c(w)}}.
$$

Note, however, that for practical reasons, we limited the number of nonzero coordinates of each vector to 1,000, taking the tfidf of the thousand most frequent words in each category. This compromise has been useful since the number of distinct words increases gradually as we move towards the top of the hierarchy, and since every page is owned by category ``\emph{Article}'', this also stands for every word and thus the vector of that category would have 339,679 nontrivial dimensions, which would make calculations too heavy for our equipment.

Having obtained the vectors of all categories, we define the \emph {weight of semantic relatedness} of any inclusion of categories $c_i\to c_j$ by
$$
p(c_i\to c_j)=\langle \vec{c_i},\vec{c_j}\rangle,
$$
where $ \langle \kern1pt. \kern1pt, \kern1pt. \kern1pt \rangle $ is the Euclidean scalar product of two vectors. Since all vectors are unitary, we have $ \mathrm {Im} (p) \subset [0,1] $. Similarly, for membership of pages to categories, we define:
$$
p(d_i\to c_j)=\langle \vec{d_i},\vec{c_j}\rangle.
$$

Let $ \mathcal {W} _p $ be the weighted Wikipedia digraph (whose vertices are pages and categories, edges are memberships of pages and inclusions of categories, and the weight is defined above). Various algorithms exist to obtain minimum weight spanning trees, the most famous being those of Kruskal and Prim. Unfortunately, these algorithms apply only to undirected graphs.

As for digraphs, there is not always a spanning tree respecting the orientation of the graph. For example, the graph

{\centering\begin{tikzpicture}
\tikzstyle{every node}=[draw]
\begin{scope}[shape=circle,minimum size=.15cm]
\node (1) at (0,1) {};
\node (2) at (1,1) {};
\node (3) at (0,0) {};
\node (4) at (1,0) {};
\end{scope}
\draw[->] (1) -- (2);
\draw[->] (4) -- (2);
\draw[->] (1) -- (3);
\draw[->] (4) -- (3);
\end{tikzpicture}

}
\noindent has no path of length $> 1$, even less a spanning tree.

However, the situation is different for the special case of graphs with a specific property: the existence of a vertex from which all others can be reached by directed paths. In this case, there is always a (directed) spanning tree. 

{\small \medskip

This is easily demonstrated through the usual method of progressive cycle removal. Indeed, let there be a cycle $abcdef$. From our hypothesis, there is a vertex $ s $ from which one can reach all vertices, so we are necessarily in one of the following three cases:

\kern-6pt

\begin{center}\scalebox{.8}{\begin{tikzpicture}
\tikzstyle{every node}=[draw]
\begin{scope}[shape=circle,minimum size=.4cm]
\node (1) at (0,0.75) {$a$};
\node (2) at (0.5,1.5) {\phantom{a}\llap{\,\,\,\smash{$b$}}};
\node (3) at (1.5,1.5) {$c$};
\node (4) at (2,0.75) {\phantom{a}\llap{\,\smash{$d$}}};
\node (5) at (1.5,0) {$e$};
\node (6) at (0.5,0) {\phantom{a}\llap{\,\,\,\smash{$f$}}};
\node (s) at (2,3) {$s$};
\end{scope}
\draw[->] (1) -- (2);
\draw[->] (2) -- (3);
\draw[->] (3) -- (4);
\draw[->] (4) -- (5);
\draw[->] (5) -- (6);
\draw[->] (6) -- (1);
\draw[->,dashed] (s) .. controls (0,3.5) and (-2,1) .. (6);
\draw[->,dashed] (s) .. controls (0,2.5) and (-0.5,1.5) .. (1);
\draw[->,dashed] (s) -- (2);
\draw[->,dashed] (s) -- (3);
\draw[->,dashed] (s) -- (4);
\draw[->,dashed] (s) .. controls (3,3) and (4,1) .. (5);
\end{tikzpicture}\kern0cm
\begin{tikzpicture}
\tikzstyle{every node}=[draw]
\begin{scope}[shape=circle,minimum size=.4cm]
\node (1) at (0,0.75) {$a$};
\node (2) at (0.5,1.5) {\phantom{a}\llap{\,\,\,\smash{$b$}}};
\node (3) at (1.5,1.5) {$c$};
\node (4) at (2,0.75) {\phantom{a}\llap{\,\smash{$d$}}};
\node (5) at (1.5,0) {$e$};
\node (6) at (0.5,0) {\phantom{a}\llap{\,\,\,\smash{$f$}}};
\node (s) at (2,3) {$s$};
\end{scope}
\draw[->] (1) -- (2);
\draw[->] (2) -- (3);
\draw[->] (3) -- (4);
\draw[->] (4) -- (5);
\draw[->] (5) -- (6);
\draw[->] (6) -- (1);
\draw[->,dashed] (s) .. controls (0,2.5) and (-0.5,1.5) .. (1);
\draw[->,dashed] (s) -- (3);
\end{tikzpicture}\kern1cm
\begin{tikzpicture}
\tikzstyle{every node}=[draw]
\begin{scope}[shape=circle,minimum size=.4cm]
\node (1) at (0,0.75) {$a$};
\node (2) at (0.5,1.5) {\phantom{a}\llap{\,\,\,\smash{$b$}}};
\node (3) at (1.5,1.5) {$c$};
\node (4) at (2,0.75) {\phantom{a}\llap{\,\smash{$d$}}};
\node (5) at (1.5,0) {$e$};
\node (6) at (0.5,0) {\phantom{a}\llap{\,\,\,\smash{$f$}}};
\node (s) at (2,3) {$s$};
\end{scope}
\draw[->] (1) -- (2);
\draw[->] (2) -- (3);
\draw[->] (3) -- (4);
\draw[->] (4) -- (5);
\draw[->] (5) -- (6);
\draw[->] (6) -- (1);
\draw[->,dashed] (s) -- (2);
\end{tikzpicture}
\rule{1.5cm}{0pt}}\end{center}

\kern-6pt

\noindent where dashed arrows represent directed paths. In the first case, all vertices of the cycle are reachable from \emph {outside} the cycle. Then we can remove the entire cycle, since all of its vertices are already reached and part of the tree. In the second case, part of the cycle is essential to reach vertices, for example, $ b $ is reached only when going through~$ a $. In this case we simply remove the edges of the cycle which are last before vertices reached from outside the cycle (in the example, this concerns the edges $ bc $ and $ fa $). Finally, in the third case, a single vertex is reached by an outside path. In this case we remove the last edge before the vertex, in our case: $ ab $. Doing this we can ``break'' all cycles while keeping the (weak) connectivity of the graph, and thus obtain a spanning tree. This proof of existence is not an efficient method for actually finding the spanning tree, since it provides us with no way of detecting cycles.

}

\medskip

One way to obtain such a spanning tree is by Chu-Liu \& Edmonds' algorithm \cite [p.~113-119]{chuliu}, published for the first time in 1965. This semi-linear algorithm returns a minimum weight forest of trees covering the digraph. The orientation of these trees (which is compatible with that of the graph) follows a very strict principle: each tree has a vertex (which we call ``root'') from which all other vertices are reachable by directed paths.

In the case of a graph that already has a vertex with such a property, the forest becomes a single tree, and we get a true \emph{directed minimal spanning tree} of the graph.

If we bothered to delve into such graph theory depths, it is because the Wikipedia category graph is indeed of this kind: when we reverse the membership/inclusion relations, then every page is reachable from the root, which is category ``\emph{Article}.''

\begin{figure}[t]
\resizebox{\textwidth}{!}{\includegraphics{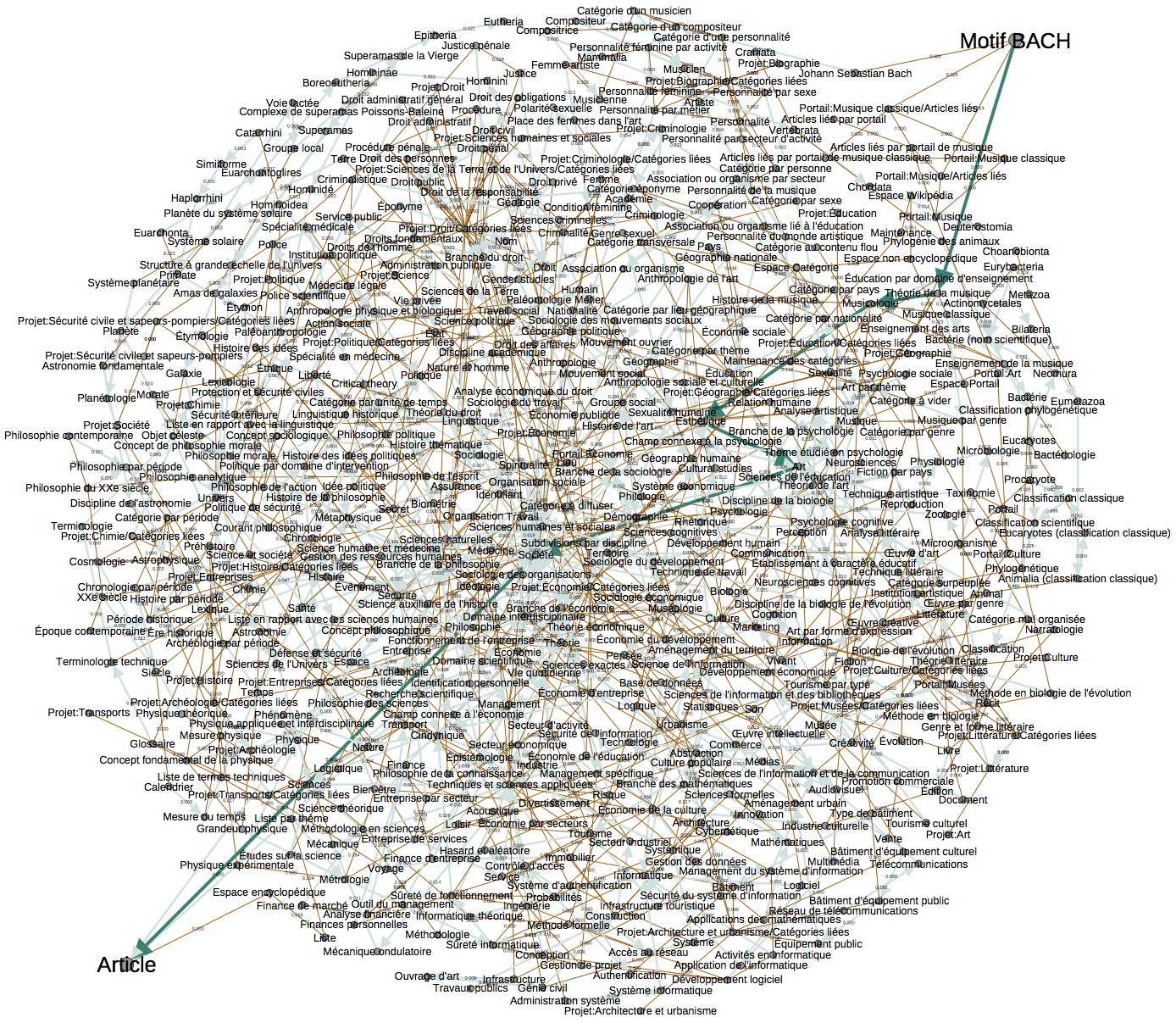}}
\caption{Graph of categories located between page ``Motif BACH'' and the top of the hierarchy (category ``\emph{Article}''). \label {fig1}}
\end {figure}

Here is an example of the results of this \emph {arborification}\footnote {The term ``arborification,'' meaning ``transformation into a tree'' while not frequent, is nevertheless used in some knowledge domains: pure mathematics (arborification of multiple divergent series) Latin literature (arborification of Baucis in a lime tree and Philemon in an oak, in Ovid, \emph {Metamorphoses} VII, 714 - 719) and (\char"21\char"21) in the interpretation of dreams. Ironically, the word ``arborification'' does not appear in French Wikipedia, whose search engine suggests ``aridification'' instead (indeed, the Levenshtein distance between the two terms is only of~3 units).}. We took the page ``\emph{Motif BACH}''\footnote {It is about the musical theme B flat, A, C, B natural, that Bach inserted in the \emph {Art of the fugue}. Other composers have used the same theme to honor Bach and so it has become a leitmotiv of the last four centuries.} and calculated the lattice which separates it from the category ``\emph{Article}.'' This lattice contains 532 categories and 1,170 category inclusions (plus the 4 edges representing the membership of ``\emph{Motif BACH}'' to the four categories ``\emph{Johann Sebastian Bach},'' ``\emph{Portail Musique classique/Articles liés},'' ``\emph{Portail Musique/Articles liés},'' ``\emph{Théorie de la musique}''). In Fig.~\ref {fig1}, the reader can see a representation of this subgraph of the Wikipedia graph. The edges belonging to the arborification of Wikipedia are drawn in thick light blue lines, those belonging to the graph but not to the tree, in thin brown lines, and finally the path connecting ``\emph{Motif Bach}'' to ``\emph{Article},'' in dark green. Note that the algorithm has chosen ``\emph{Théorie de la musique}'' rather than the category carrying the composer's name as the semantically most related category. This can be explained by the fact that the page ``\emph{Motif Bach}'' focuses rather on works by other composers than on those by Bach, despite the obvious trap that the page title and the category ``\emph{Johann Sebastian Bach}'' both contain the composer's name.

Remember, however, that our goal is not to find the most appropriate taxonomy, i.e., the most relevant path from a page to the top. To get it, it would be wiser to use a shortest path algorithm, such as the one by Dijk\-stra. This has already been proposed in \cite {Collin:2010}, but for the metric of the number of edges; in our case we would rather use our weighting of the graph. 

\section{Stratified Tfidf}

We will use the page ancestors in the minimal spanning tree to update tfidf values of words in the page vectors. Recall the definition of a Wikipedia page vector, given in Section~\ref{vec}:
$$
\vec{d}:=\frac{\sum_{w\in d}t_d(w)\cdot \vec{w}}{\sqrt{\sum_{w\in d}t^2_d(w)}},
$$
where $ t_d (w) $ is the tfidf of word $ w $ for document $ d $. We will change function  $ t_d$, so that it takes the ancestor categories of page $ d $ into account.

We were motivated by the following problem: sometimes one finds words with relatively high tfidfs in pages belonging to different contexts (e.g., in a slightly caricatural case: the word with highest tfidf of page ``\emph{Roman noir}'' is\ldots\ ``\emph{mai-juin},'' (=~May-June) because in the bibliography, a book published in May-June 2001 is cited thrice). Our method aims to reduce the influence of out-of-context words, the presence of which can decrease relevance of concept vectors.

For this, we chose to give more importance to the tfidf of words being in the ``right context.'' It is reasonable to assume that the ``right context'' is one that is consistent with the categories to which the page belongs, and, in particular, with the main category and other ancestors obtained by taking the tree path that leads to the root. 
But according to our definition, words have tfidf values in categories. So we just use these tfidf values to boost words that ``survive'' when moving to categories, that is to say, words that appear in both the page vector and in category vectors.

To do this, we replace $t_d$ by $t'_d $, defined as follows:
$$
t'_d(w)=t_d(w)+\sum_{i\geq 0}\lambda_i t_{\pi^i(d)}(w)
$$
where $ \pi $ is the operation replacing an node of the tree by its parent, and $ \pi ^ i $ is its \mbox {$ i $th} iteration  (for practical reasons, we restricted ourselves to $ i \leq3$). The coefficients $ \lambda_i $s must form a decreasing sequence, and we tested the following three cases: $\lambda_i=\frac{1}{2^i}$, $\lambda_i=\frac{1}{2^{i-1}\cdot10}$, $\lambda_i=1$.

We call this new tfidf \emph{stratified} because we move upwards through the different strata to provide a semantic consolidation of the page stratum, which is the lowest.

We followed the same modus operandi as in section~\ref {svm} and obtained the following results:
\setlength\dashlinedash{1pt}
\setlength\dashlinegap{1pt}\begin{center}
\small\begin{tabular}{|c|c|c|c|}\hline
Strata $\lambda_i$&Dim. of concept subspace&Number of SVs&Precision\\\hline
$\frac{1}{2},\frac14,\frac18$&121,114&1,280&\textbf{88.62\%}\\
$\frac{1}{10},\frac{1}{20},\frac{1}{40}$&119,918&1,298&88.50\%\\
$1,1,1$&119,593&1,347&87.90\%\\\hdashline
(without strates)&103,291&665&70.08\%\\
\hline
\end{tabular}

\end {center}
which are significantly better than the conventional ESA for the same corpus (accuracy of 70.08\%), This confirms our approach. Note that between standard and stratified ESA, the increased accuracy is consistent with the increase in the size of the subspace of concepts and in the number of support vectors (which practically doubled while the dimension of the subspace of concept vectors increased by only 20\%). On the other hand, within the group of our three tests, we maintain the correlation between accuracy and dimension of the concept subspace but the correlation between precision and number of support vectors has been reversed. We also note that the variations between the first two types of strata are small, and that the caricatural choice of constant $ \lambda_i= 1$ (which means that we give equal importance to the categories and to the page itself) has degraded performance, but only slightly. Finally, the last observation: needing 121,114 concepts (= pages), means that we use 94\% of Wikipedia pages, while the nonstratified method uses only 80\% of them, which shows that our method was able to benefit more from the semantic richness of the corpus, mobilizing a larger number of pages.

\section {Conclusion and Hints for Further Research}

A study of the Wikipedia category graph has revealed that one can extract a directed spanning tree, whose root is the category ``\emph{Article},'' the hierarchical summit of the corpus. By applying Chu-Liu \& Edmonds' algorithm, we obtained a minimum spanning tree---minimum with respect to the metric induced from explicit semantic analysis and adapted to categories by the notion of categorical tfidf. In this tree, every page and every category have a unique list of ancestors. We use this list to define a stratified tfidf, whose strata are the the values of the tfidfs a given term takes in the various ancestor categories.

We evaluated our method by a text classification task: stratified tfidf allowed us to improve the classification accuracy by 18\%.

Here are some hints for research to be done:
\begin {enumerate}
\item reduce the size of arborified Wikipedia by removing the intermediate, nondiscriminative categories;
\item quantify the relevance of the hierarchy obtained (both by frequentist methods and by comparing, for example, with the taxonomy of WordNet concepts), find its weak areas, compare with versions of Wikipedia in other languages;
\item examine the impact of the restriction on the maximum size of vectors for high-level categories (in this article we placed an artificial limit of 1,000 nonzero values---when the leaves of categories are becoming more and more, this value can prove to be binding);
\item aggregate our stratified measure with collocational and ontological components, as in \cite {Haralambous:2011wm};
\item propose the notion of the ``most relevant category'' to Wikipedia users and use their answers to improve the system;
\item propose an alternative measure, based on links between pages (or categories), proportional to the number of links (or link paths) between pages and inversely proportional to the length of these paths. Properly define this measure and compare it to ESA (which uses the number of links between pages to filter Wikipedia, but does not include it into semantic relatedness calculations);
\item in this article as well as in the hints above, the amount of data processed is becoming increasingly monstrous (especially when applied to English Wikipedia). Find the best dimensionality reduction methods to make them implementable;
\item and, more generally, explore the applications of graph theory to the formidable mathematical-linguistic objects that are the different graphs extracted from Wikipedia.
\end {enumerate}

\subsection* {Acknowledgements, Technical Details, Availability} {\small
The authors thank Julie Soulas, tireless student in the CS Department of Télécom Bretagne, who helped in building and debugging the system, as well as Sébastien Bigaret, the  administrator of the server Eole of the Lab-STICC laboratory, who allowed them to perform  calculations under good conditions.

All programs were coded in Perl (with one exception: for a product of very large matrices,  C language allowed better memory management). Chu-Liu \& Edmonds' algorithm has been implemented by Petr Pajas. The graphs in Fig.~\ref{fig0} are done in~R. The graph in Fig.~\ref{fig1} has been layouted by Gephi, using the Fruchterman-Reingold algorithm. For stemming, the flemm tool has been used. SVM classification was performed using the e1071 library in~R.
All resources obtained through this project are placed in the public domain and will be available online (please contact the authors).

}
\bibliographystyle{apalike}
\bibliography{paper}

\end{document}